\documentclass[sigconf]{acmart}
\usepackage{multirow}
\usepackage{url}
\usepackage{footmisc}
\usepackage{hyperref}
\usepackage{xcolor}
\AtBeginDocument{%
  \providecommand\BibTeX{{%
    \normalfont B\kern-0.5em{\scshape i\kern-0.25em b}\kern-0.8em\TeX}}}

\copyrightyear{2020}
\acmYear{2020}
\setcopyright{acmcopyright}
\acmConference[CIKM '20]{Proceedings of the 29th ACM International Conference on Information and Knowledge Management}{October 19--23, 2020}{Virtual Event, Ireland}
\acmBooktitle{Proceedings of the 29th ACM International Conference on Information and Knowledge Management (CIKM '20), October 19--23, 2020, Virtual Event, Ireland}
\acmPrice{15.00}
\acmDOI{10.1145/3340531.3417445}
\acmISBN{978-1-4503-6859-9/20/10}

\begin{document}

\title{Large scale long-tailed product recognition system at Alibaba}

\author{Xiangzeng Zhou, Pan Pan, Yun Zheng, Yinghui Xu, Rong Jin}
\affiliation{%
  \institution{Machine Intelligence Technology Lab, Damo Academy \\Alibaba Group, Hangzhou, China}
  }
\email{xiangzeng.zxz,panpan.pp,zhengyun.zy@alibaba-inc.com }
\email{renji.xyh@taobao.com,jinrong.jr@alibaba-inc.com}

\renewcommand{\shortauthors}{Zhou and Pan, et al.}

\begin{abstract}\label{sec:abs}
A practical large scale product recognition system suffers from the phenomenon of long-tailed imbalanced training data under the E-commercial circumstance at Alibaba. Besides product images at Alibaba, plenty of image related side information (e.g. title, tags) reveal rich semantic information about images. 
Prior works mainly focus on addressing the long tail problem in visual perspective only, but lack of consideration of leveraging the side information.
In this paper, we present a novel side information based large scale visual recognition co-training~(SICoT) system to deal with the long tail problem by leveraging the image related side information.
In the proposed co-training system, we firstly introduce a bilinear word attention module aiming to construct a semantic embedding over the noisy side information. A visual feature and semantic embedding co-training scheme is then designed to transfer knowledge from classes with abundant training data (head classes) to classes with few training data (tail classes) in an end-to-end fashion.
Extensive experiments on four challenging large scale datasets, whose numbers of classes range from one thousand to one million, demonstrate the scalable effectiveness of the proposed SICoT system in alleviating the long tail problem. In the visual search platform Pailitao\footnote{http://www.pailitao.com} at Alibaba, we settle a practical large scale product recognition application driven by the proposed SICoT system, and achieve a significant gain of unique visitor~(UV) conversion rate.

\begin{figure}[t]
  \centering
  \centerline{\includegraphics[width=8.6cm]{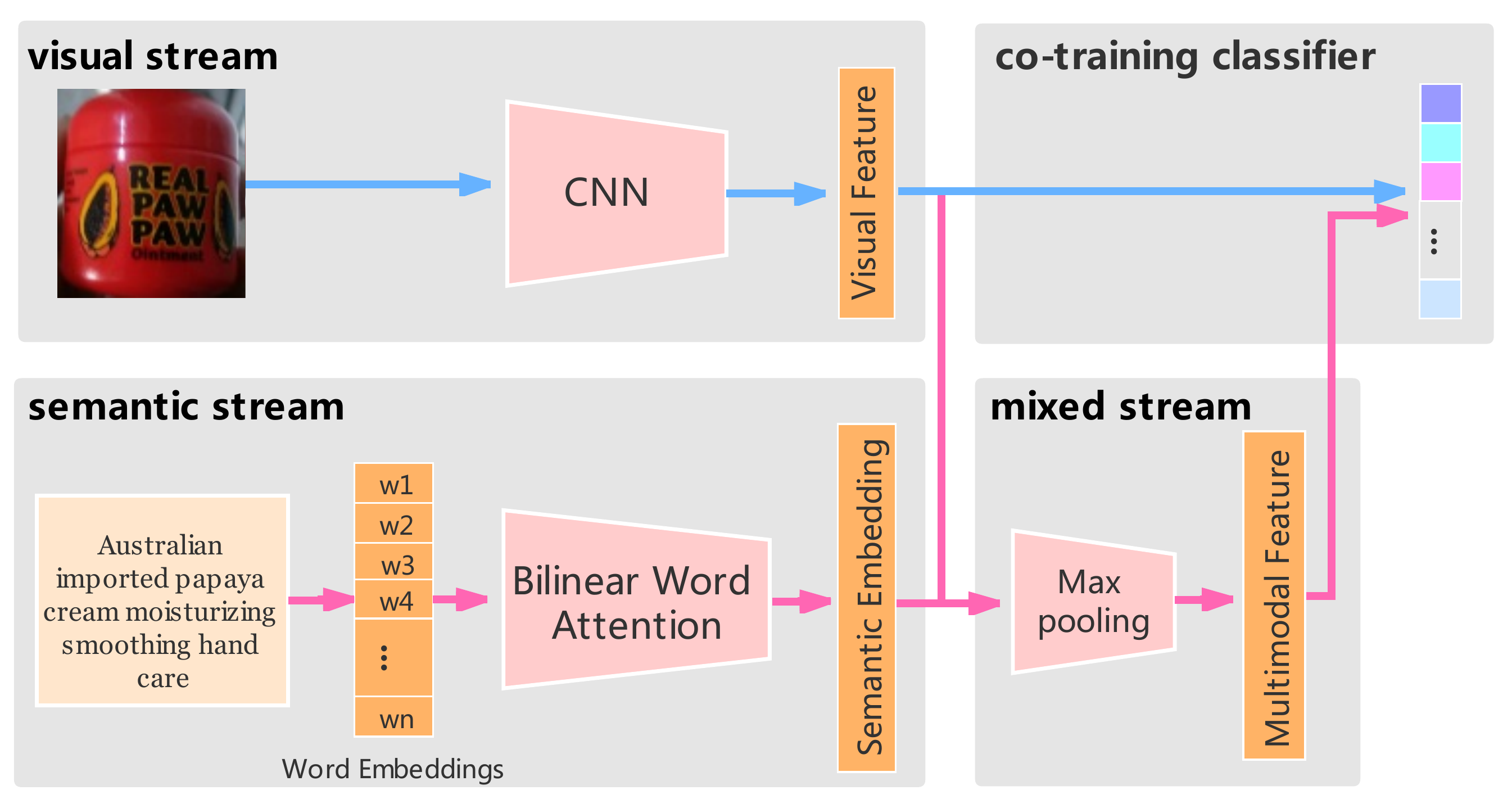}}
\caption{The overall architecture of our proposed side information based co-training~(SICoT) system. The system contains four streams: (i) The visual stream is a visual feature extractor using a convolutional neural network. (ii) The semantic stream, which consists of a word2vec module and a proposed bilinear word attention module, aims to learn a semantic embedding from the noisy side information. (iii) The mixed stream takes charge of generating a multimodal feature. (iv) The shared classifier is co-trained by the visual stream and the mixed stream. }
\label{fig:cotrain}
\end{figure}
\end{abstract}



\maketitle

\section{Introduction}\label{sec:intro}
Recent years have witnessed the remarkable progresses of wielding deep learning models in visual recognition task.
With the aid of deep learning techniques, nowadays it is practicable to establish an industrial large scale visual recognition application based on huge volume of image data. Compared to the quantity of products and image data under the E-commercial circumstance at Alibaba, many popular so-called large scale visual recognition datasets, like ImageNet~\cite{deng2009imagenet}, WebVision2.0~\cite{WebVision2017}, iMaterialist Product 2019~\cite{imat2019} and Open Images V4~\cite{kuznetsova2018openimage}, appear to be relatively small scale. Even though some datasets have reached several millions of training image data, the quantity of categories only ranges from hundreds to thousands.

On the basis of abundant image data of great value in the E-commercial scenario and powerful computing resources at Alibaba, it is still great challenging to establish a truly practical large scale visual recognition application. And these challenges are roughly reflected in following three aspects:

\textbf{Enormous quantity of classes and images:}
There are about tens of million of daily active products and billions of image data in the marketplace of Alibaba, which covers categories of clothing, shoe, bag, cosmetic, drink, snack and toy in general. The huge quantity of product classes and images brings difficulties to both the training and deployment of a large scale visual recognition model. For example, when the number of classes reaches about one million, the size of the last fully connected~(FC) layer will exceed the maximum memory of a single block of Nvidia-V100-32G GPU. This requires a new training paradigm capable of training a huge FC layer in a distributed manner.

\textbf{Scarce and noisy annotation:}
Unlike those well compiled small scale datasets (e.g. ImageNet~\cite{deng2009imagenet}), it is impracticable to manually annotate the daily growing enormous quantity of image data in the E-commercial scenario. Training images with high-quality annotations are scarce and insufficient to build a practical large scale visual recognition system. Although there are lots of annotations provided by sellers or customers in the marketplace of Alibaba, it is of great difficulty to use these weakly and noisy annotations to assist in training a visual recognition model.

\textbf{Long-tailed distribution of training data:}
The phenomenon of long-tailed imbalanced training data naturally occurs under the E-commercial circumstance. In the marketplace of Alibaba, a large amount of new arrival products emerge everyday, meanwhile, quite a large portion of products are low sales or even zero sale. The difficulty to acquire sufficient training images for these products challenges the performance of a large scale visual recognition application.

It is known that, without any special treatment of the classes with insufficient training data (tail classes), the classification boundary of a recognition model inclines toward those classes with abundant training data (head classes).
At Alibaba, there are abundant image related data or side information, such as short titles and long text descriptions, coming from various sellers or customers. These side information that containing rich, yet weak and noisy annotations reveal underlying similarity among images and classes from a different perspective.

However, prior works~\cite{he08,Maciejewski11,Tang09,Zhou06,khan2017cost,castro2013novel,dong19,dong17,xiao17} mainly focus on addressing the long tail problem in visual perspective only, but lack of consideration of leveraging the side information. On the other hand, most works~\cite{joulin2016learning,corbiere2017leveraging,ma2019and} take advantage of the side information as a kind of weakly supervision in general, and are not meant to address the long tail problem. Inspired by the work of transferring knowledge or borrowing training examples between similar classes~\cite{lim2011transfer} in detection task,
we attempt to address the problem of long-tailed distributed training data in the task of large scale product recognition by leveraging the noisy side information in this paper.
Considering the following two facts observed on the data of the marketplace at Alibaba, the usage of image related side information has great potential to alleviate the long tail problem.
a)~Unlike the extreme imbalanced distribution of image data, the distribution of words from image related titles is relatively much balanced.
b)~About 12\% words out of the entire vocabulary are shared between the head classes and the tail classes.

\begin{figure}[t]
  \centering
  \centerline{\includegraphics[width=8.5cm]{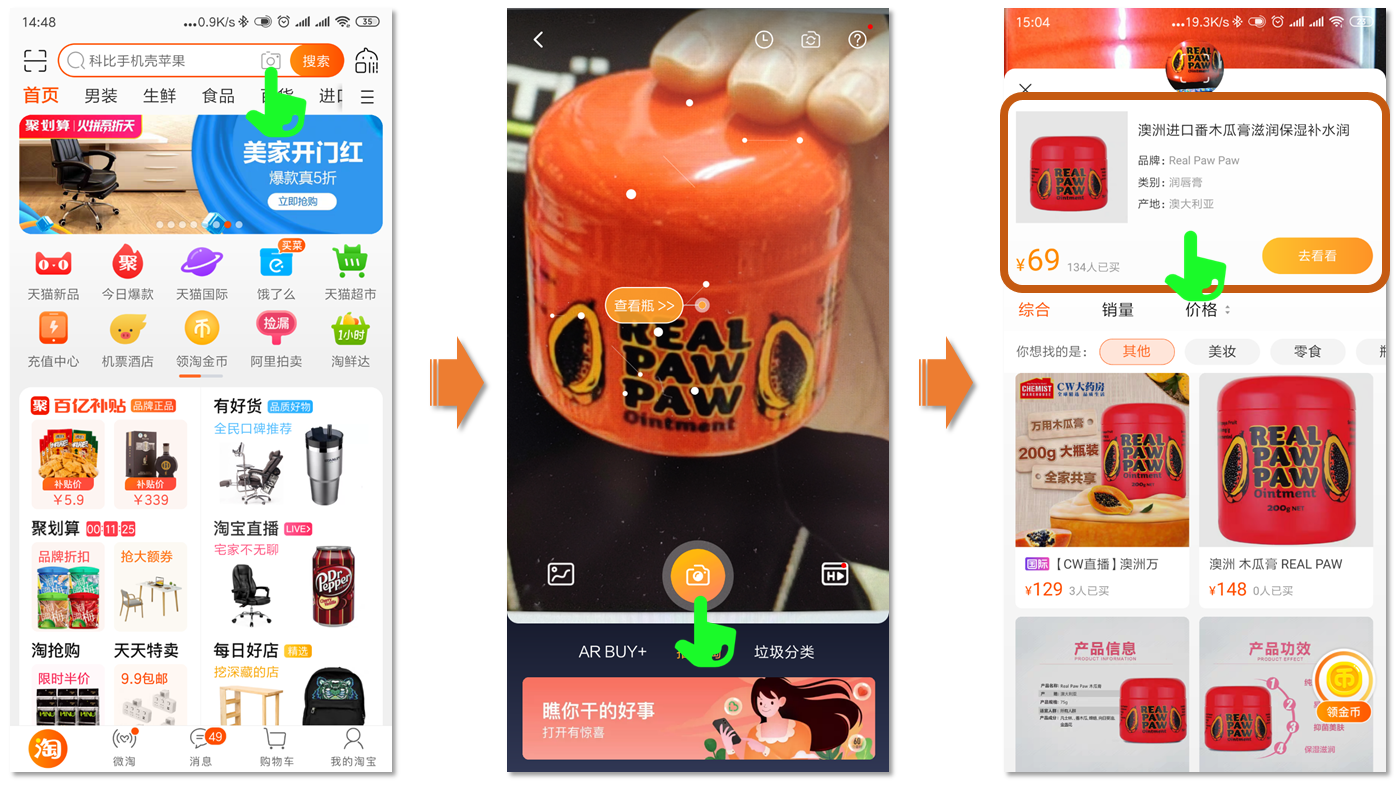}}
\caption[]{The scenario of large scale SKU level product recognition service on Pailitao at Alibaba: by taking a picture, Pailitao identifies the product with its short title and several associated tags in real time shown at the top of the page.}
\label{fig:demo}
\end{figure}

In this paper, we propose a novel side information based visual recognition co-training~(SICoT) system, as shown in Fig.~\ref{fig:cotrain}, which aims to deal with the long tail problem in a large scale product classification task. Moreover, we have launched a SKU level product recognition service driven by the proposed SICoT system in the visual search platform Pailitao~\cite{plt,Zhang:2018} at Alibaba, as shown in Fig.~\ref{fig:demo}. We conclude our contributions as following:

1) We introduce a bilinear word attention module to distinguish important words from the noisy side information of image short titles, followed by constructing a semantic embedding as a kind of distilled knowledge of the side information.

2) Considering the long tail problem in a large scale product recognition task, we propose a novel visual feature and semantic embedding co-training~(SICoT) system to help transfer knowledge from the head classes to the tail classes in an end-to-end way. This co-training system aims to perform transfer learning across the head and the tail classes by deeply involving the side information in both feature learning and classifier training.

3) Extensive experiments on both open large scale datasets and our organized huge scale SKU level product datasets demonstrate the scalable effectiveness of the proposed side information based co-training system in relieving the long tail problem.

\section{Related Work}\label{sec:work}
Taking the issue of long-tailed distributed training data into account, it still remains very challenging problems on how to establish a practical large scale product recognition system in the E-commercial scenario at Alibaba. Some prior works about addressing the long tail problem, taking advantaging of side information and large scale product recognition are roughly summarised in the following aspects.
\begin{figure*}[t]
  \centering
  \centerline{\includegraphics[width=16cm]{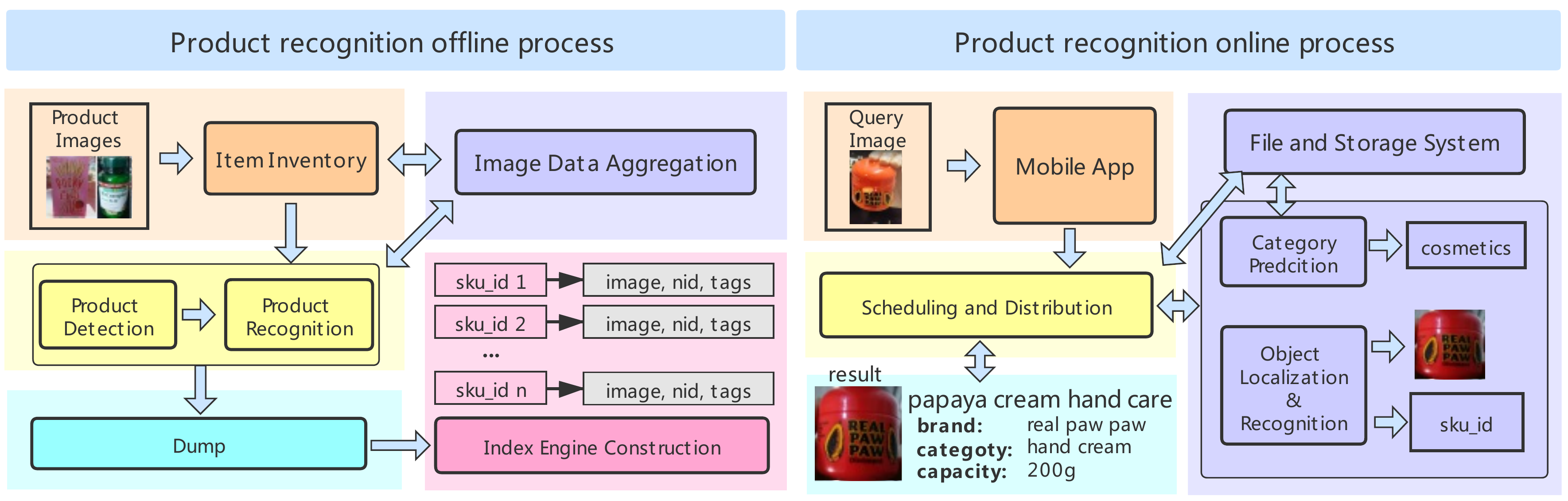}}
\caption[]{Overview of the overall product recognition architecture settled in Pailitao.}
\label{fig:arch}
\end{figure*}

\textbf{Imbalanced learning:}
To alleviate the problem of long-tailed distributed training data, many traditional approaches have been extensively studied in the past~\cite{he08,Maciejewski11,Tang09,Zhou06,khan2017cost,castro2013novel,dong19,dong17,xiao17}. Re-sampling methods~\cite{he08,Maciejewski11} aim to balance the numbers of training samples between multiple classes by under-sampling the head classes or over-sampling the tail classes. However, these methods often lead to removal of important samples or introduction of meaningless duplicated samples.
Cost-sensitive methods~\cite{Tang09,Zhou06} try to make the standard classifiers more sensitive to the head classes by imposing higher misclassification cost to the head classes than to the tail classes.
Most recently, deep neural networks are widely applied to performing imbalanced learning~\cite{Zhou06,khan2017cost,castro2013novel,xiao17,dong17,dong19}. Apart from the conventional cross entropy loss, several new objective loss function, like range loss~\cite{xiao17}, class rectification loss~\cite{dong17} and cluster-based large margin local embedding~\cite{dong19} are proposed to address the long tail problem by rectifying the classification boundaries dominated by the head classes.
The works mentioned above devote major effort to addressing the long tail problem merely from the visual aspect. Considering that the side information can provide rich semantic information about image, in this paper the side information are involved in helping tackle the long tail problem.

\textbf{Weakly supervised learning:}
In many popular datasets and practical scenarios, lots of auxiliary data or side information associated with images is provided, such as image titles and long text descriptions of in WebVision2.0~\cite{WebVision2017}, wordnet in ImageNet~\cite{deng2009imagenet}
and so on. These side information normally come from heterogeneous data sources via web search, and naturally contain a lot of noise. In most of prior works, the noisy side information is mainly taken as a kind of weakly supervision in coordination with other kinds of learning tasks~\cite{joulin2016learning,corbiere2017leveraging,ma2019and}. However, especially in a classification task, taking advantage of the side information as supervision in the one-hot fashion may not fully exploit the knowledge and be somewhat shallow. Besides, the usage of the side information is these works are not meant to address the long tail problem. In our proposed SICoT system, we explore a novel fashion to handle the long tail problem by taking advantage of the side information.

\textbf{Knowledge distill and transfer:}
The basic principle of transfer learning are also introduced to transfer knowledge or even borrow training data from the head classes to the tail classes~\cite{salakhutdinov2011learning,lim2011transfer,zhu2014capturing}.
A series of works~\cite{Vladimir09,Vapnik15,Yang18} present a learning using privileged information~(LUPI) framework to transfer knowledge (e.g. similarity or margin) across multiple models which usually learned in different modalities. A teacher-teach-student scheme presented in the LUPI framework provides a relatively deeper way to use the side information to assist the original visual recognition task. Extending to this, \cite{Lopez-Paz16,wang19} unify the LUPI framework and the knowledge distillation paradigm~\cite{Hinton15-dist} into a generalized distillation framework. These works provide a possible way to use the side information as a teacher model to help a visual recognition task~(student model) in a teacher-teach-student or distillation scheme. However, the teacher-teach-student scheme requires the teacher model to be learned beforehand. And this two-stage approach is hard to be optimized globally. This scheme also implicitly demands the teacher model to be more powerful than the student model. In our proposed SICoT system, the side information participate into the visual classification task in conjunction with the visual feature in an end-to-end way. Moreover, no prior assumptions and constraints are made on the side information.

\textbf{Large scale product recognition:}
Taking the applied techniques into account, Trax~\cite{trax,cohen2019method} and MalongTech~\cite{malong} have devoted their efforts to establish practical large scale visual recognition applications, especially the SKU~(stock keeping unit~\footnote{https://en.wikipedia.org/wiki/Stock\_keeping\_unit}) level product visual recognition. MalongTech hosts a SKU level product dataset iMaterialist product 2019 and a corresponding competition in conjunction with FGVC6 workshop of CVPR2019~\cite{imat2019}. However, the iMaterialist product 2019 dataset covers only two thousands of product SKUs, and only provides image data without any side information for extending research. At Alibaba, in order to establish a scalable product recognition system, we organize a large scale SKU level product dataset that consists of about 60 million real-shot product images covering 1 million SKUs with the aid of the visual search engine Pailitao~\cite{plt,Zhang:2018}. Apart from images, the dataset contains abundant yet noisy side information (e.g. image titles, long descriptions and tags) provided by various sellers or customers in the marketplace of Alibaba. On the basis of the large scale SKU level product dataset, we settle a 30 million products recognition service driven by the proposed SICoT system in Pailitao. To our knowledge, this is the largest scale product recognition application in the E-commercial scenario so far.

\section{Approach}\label{sec:approach}
In this section, we elaborate our proposed side information based co-training~(SICoT) system in a large scale visual recognition task over long-tailed distributed training data. In Sec.~\ref{sec:arch}, we firstly illustrate the overall product recognition process on the basis of the visual search service Pailitao~\cite{plt,Zhang:2018}. Considering that the side information of image titles from heterogeneous resources are often noisy, we then propose a bilinear word attention network in Sec.~\ref{sec:uce} to distinguish the important words from the noisy side information. Subsequently, a detailed illustration of the side information based co-training system~(SICoT) is given in Sec.~\ref{sec:cotrain}. The SICoT system aims to leverage visual features and semantic embeddings to help transfer knowledge from the head classes to the tail classes in an end-to-end fashion.

\subsection{Product Recognition Architecture}\label{sec:arch}
The entire product recognition architecture, as shown in Fig.~\ref{fig:arch}, comprises an offline process and an online process, that following the present visual search architecture of Pailitao~\cite{plt,Zhang:2018} in general. The offline process mainly refers to the daily process of building product index using the proposed SICoT product recognition system. Unlike the index engine in the visual search service, the product index stores the SKU ids of products and corresponding product images, titles and tags for online retrieval. In the online process, the core function is a real time product recognition service in charge of predicting a SKU id for each query image. For the other modules in this architecture, we simply reuse the design of the visual search service, like category prediction and object localization. Once a query image is successfully recognized by the online service, a predicted SKU id will be obtained. By retrieving the index engine using the SKU id, the corresponding product image, title and tags will be obtained and presented to the customer, as shown in Fig.~\ref{fig:demo}.

\begin{figure}[t]
  \centering
  \centerline{\includegraphics[width=6.5cm]{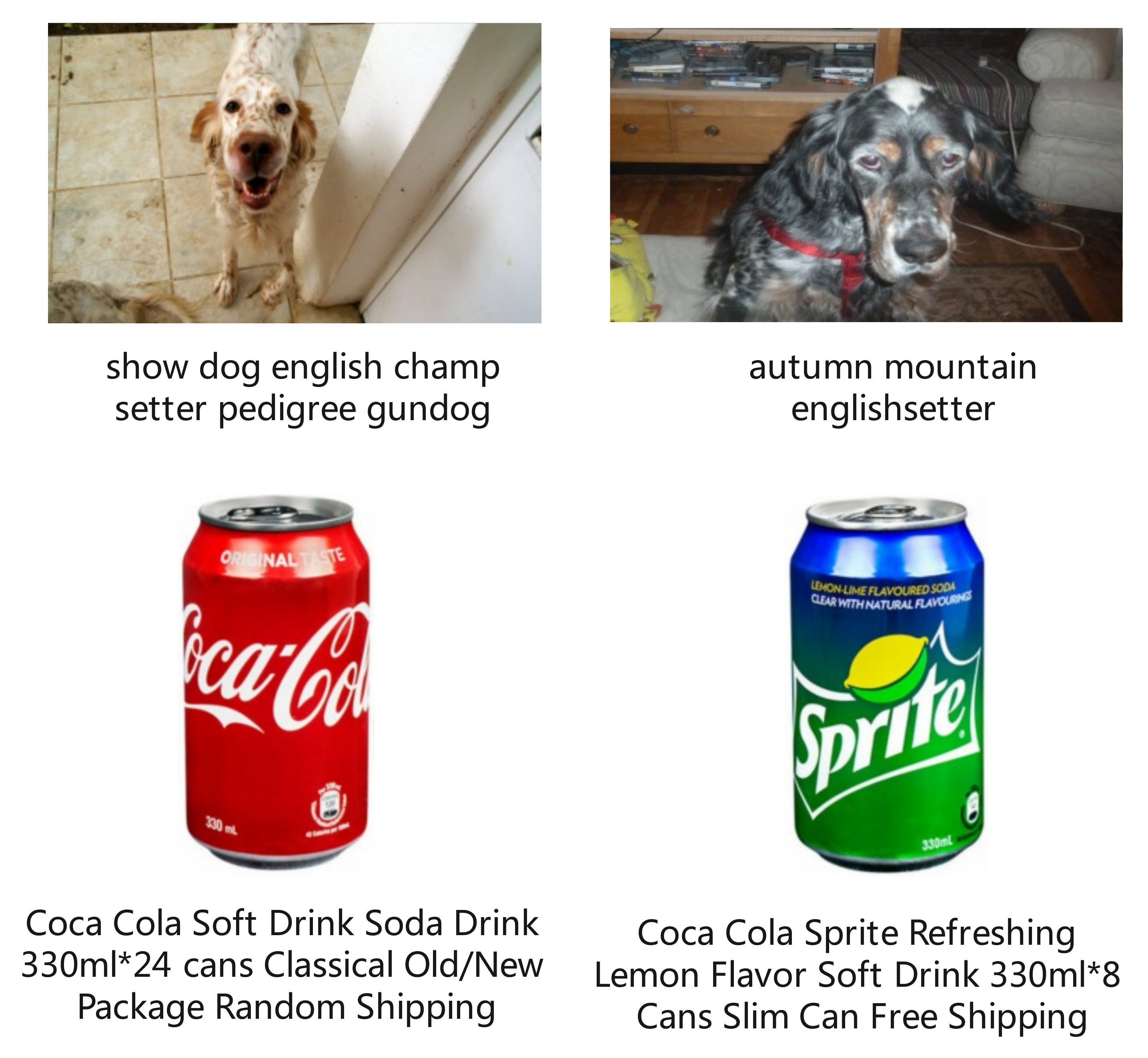}}
\caption{Two images of English setters from the Webvision2.0~\cite{WebVision2017} and two images of products from the marketplace of Alibaba. Along with each image, a short text description is also provided. These text descriptions are naturally noisy.}
\label{fig:exp-side}
\end{figure}

\subsection{Bilinear word attention network}\label{sec:uce}
\begin{figure*}[t]
  \centering
  \centerline{\includegraphics[width=18cm]{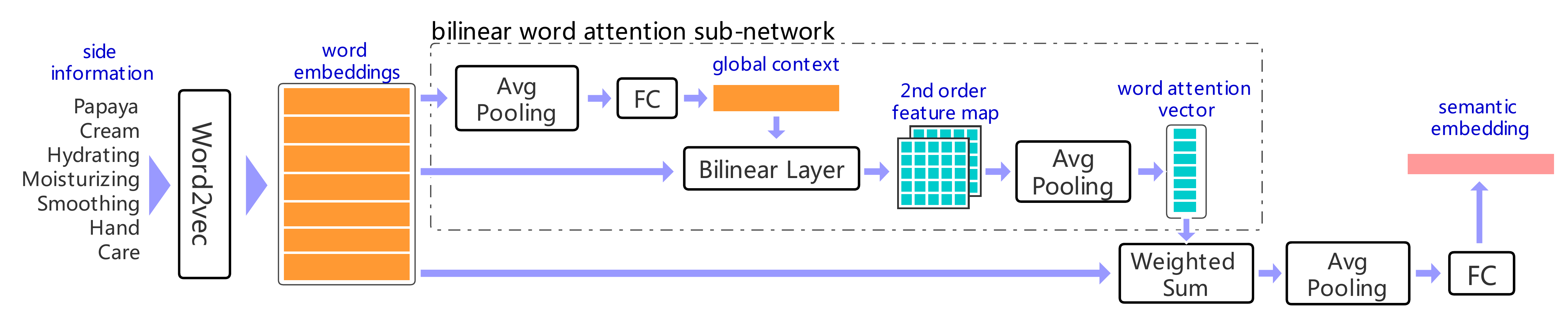}}
\caption{The semantic embedding network mainly consisting of a word2vec module and the proposed bilinear word attention module. }
\label{fig:uce}
\end{figure*}

In many practical scenarios, besides images lots of related side information (e.g. image titles) can be obtained. These image related side information may reveal underlying similarity among images and classes, so as to it has great potential to improve a visual recognition task. In order to process both the visual information of images and the image related side information in an unified framework, we propose a bilinear word attention network, as shown in Fig.~\ref{fig:uce}, to learn a semantic embedding from the noisy side information.

A conventional tokenization is firstly conducted on the side information of image titles, followed by using a word2vec model to generate a word embedding for each word after tokenization. In most of natural language processing~(NLP) related tasks, such as language translation and image captioning, the order of words in a title matters in general. However, in our experimental observations, the order of words is less important and even harmful, especially when the side information is highly noisy. Here, we simply use an average pooling operation to generate a global embedding from the word embeddings instead of using a sequential model~(e.g. recurrent neural networks). As for each word embedding, this global embedding can be regarded as a global context without consideration of the order of words in the side information.

Noise and meaningless words naturally occur in the image related side information due to the heterogeneous resources in both the marketplace of Alibaba and many open datasets. For example, as shown in Fig.~\ref{fig:exp-side}, both the text descriptions of two dogs from Webvision2.0 and the titles of two soft drinks from the marketplace contain several words less relevant to the image content~(e.g. \textit{show}, \textit{autumn}, \textit{free shipping}). Considering this issue, we propose a soft attention sub-network to evaluate the importance of each word in the side information. A bilinear operation between the global context and all word embeddings is introduced to generate a second order feature map. An average pooling operation and a nonlinear transformation are then carried out above the feature map to output a word attention vector. A semantic embedding of the entire side information is achieved by the weighted sum of the word attention vector and all word embeddings at final.

\subsection{Side information based co-training system}\label{sec:cotrain}
Given the bilinear word attention based semantic embedding presented in Sec.~\ref{sec:uce}, we propose a visual feature and semantic embedding co-training scheme in this section. Due to the absence of image related side information once a recognition model has been deployed, the proposed co-training scheme is only involved in the training stage. The scheme is designed to take the semantic embedding from the side information as a bridge to transfer knowledge from the head classes to the tail classes.
Unlike the teacher-teach-student paradigm used in LUPI~\cite{Vapnik15}, our approach makes no prior assumption about models, i.e. a teacher model should be more powerful than a student model. In fact, in our experimental observations, it is often inadequate to carry out a satisfactory classification by using the image related side information only, especially when the number of class are huge.

As illustrated in Fig.~\ref{fig:cotrain}, we show the overall architecture of our proposed co-training scheme that consisting of three streams, i.e. a visual stream, a semantic stream and a mixed stream. The visual stream is a conventional visual recognition pipeline, which comprises a common convolutional neural network as a feature extractor and a plain softmax classifier optimized by a cross entropy loss. The visual stream is the target task that we attempt to improve. The semantic stream is simply the bilinear word attention based semantic embedding sub-network, in which the word embeddings are required initialized from a pretrained model like Word2vec~\cite{mikolov2013efficient}. Noted that it is assured that the visual feature and the semantic embedding take a same dimension through a deliberate design of the network.
A mixed or multi-modal feature is then achieved by a max-pooling operation over the visual features and the semantic embeddings. Unlike a conventional  multi-task framework that consisting of multiple learning tasks driven by different objectives, the proposed co-training scheme makes both the visual features and the semantic embeddings to be learnt driven by a same classification task.
As shown in Fig.~\ref{fig:cotrain}, the visual feature $\mathbf{x}^{\mathrm{v}}$ and the multi-modal feature $\mathbf{x}^{\mathrm{m}}$ are followed by a shared co-training classifier optimized with the objective as Equ.~\ref{equ:loss}, in which the $\hat{\mathbf{y}}^{\mathrm{v}}_{i}$ and $\hat{\mathbf{y}}^{\mathrm{m}}_{i}$ are the classifier output of the visual feature $\mathbf{x}^{\mathrm{v}}_i$ and the semantic embedding $\mathbf{x}^{\mathrm{m}}_i$, respectively.

\begin{align}
  Loss & = - \frac{1}{N}(\sum_{i=1}^{N}\mathbf{y}_i \cdot \mathrm{log}( \hat{\mathbf{y}}^{\mathrm{v}}_{i} )) - \lambda \cdot \frac{1}{N}(\sum_{i=1}^{N}\mathbf{y}_i \cdot \mathrm{log}( \hat{\mathbf{y}}^{\mathrm{m}}_{i} ))
  \label{equ:loss}
\end{align}

In the proposed training scheme, the shared classifier are learnt in a co-training fashion, in which the classification boundaries are directly affected both visually and semantically. It is well known that the classification boundaries in a classification task with long-tailed imbalanced training data are easily dominated by the head classes. The proposed co-training scheme may rectify the skewed classification boundaries by introducing the semantic knowledge into the classification training. Furthermore, it can be observed in Fig.~\ref{fig:cotrain} that the visual and semantic streams are tangled in not only the classifier training part, but also the feature learning part via the gradient backward procedure. Compared to the methods of taking the side information as weakly supervision and the two stage teacher-teach-student paradigm in the LUPI framework, our proposed co-training scheme provides a deeper way to tangle the visual and semantic knowledge in an end-to-end manner.

\section{Experiments}
\label{sec:exp}
\begin{figure*}[htb]
\begin{minipage}[b]{.98\linewidth}
  \centering
  \centerline{\includegraphics[width=12.0cm]{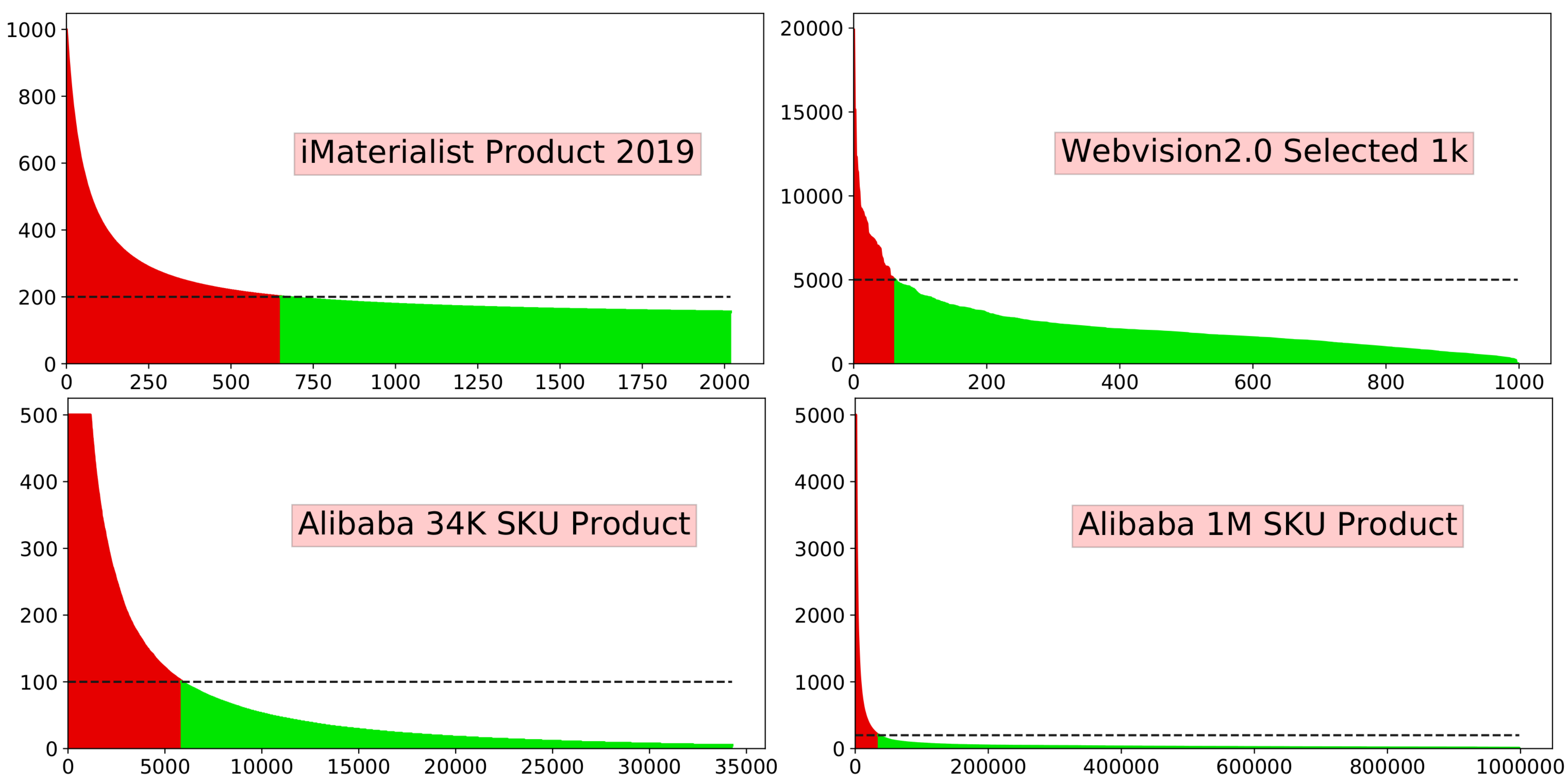}}
\end{minipage}
\caption{Long-tailed distribution of training set of the four datasets. The red and green parts in each panel represent the head and tail classes, respectively.}
\label{fig:longtail-dist}
\end{figure*}

\begin{table*}[tb]\centering
\begin{tabular}{|c|c|c|c|c|}
\hline
\multicolumn{1}{|l|}{}                                                        & {Class} & {Trainset} & {Testset} & {Vocab} \\ \hline
{iMaterialist Product 2019} & 2019           & 440438            & 9986             & 175208         \\ \hline
{Webvision2.0 Selected 1k}  & 1000           & 2230968           & 59040            & 162541         \\ \hline
{Alibaba 34K SKU Product}   & 34258          & 2305853           & 171290           & 19762          \\ \hline
{Alibaba 1M SKU Product}    & 998131         & 55776960          & 7675690          & 345656         \\ \hline
\end{tabular}
\vspace{1em}
\caption{Statistics of the four datasets, in which the ''Vocab'' means the number of words extracted from the side information of image titles after a tokenization process. }
\label{tab:ds}
\end{table*}

In this section, we evaluate our proposed side information based co-training approach on four large scale datasets that exhibiting long-tailed distribution, i.e. iMaterialist Product 2019~\cite{imat2019}, Webvision2.0 selected 1k~\cite{WebVision2017} and our proposed 34k and 1M SKU level product datasets, and demonstrate the positive effect of our approach on the long tail problem. We also report a relative gain of unique visitor (UV) conversion rate after settling our approach to the visual search application Pailitao~\cite{plt} at Alibaba.

\subsection{Datasets preparation}\label{ssec:dataset}
The WebVision2.0~\cite{WebVision2017} dataset is designed to facilitate the research on learning visual representation from noisy web data and it is attached with Google and Flickr retrieval results that containing titles and detailed text descriptions. In this paper, we extract the title of each image as the side information for experiments. We preprocess these side information by discarding the punctuation marks and $5$ percent of very frequent and very infrequent words, and throwing away those images without any side information. Eventually, one thousand classes are randomly picked out from the entire $5$ thousand of classes in Webvision2.0~\cite{WebVision2017}. iMaterialist Product 2019~\cite{imat2019} hosted by MalongTech~\cite{malong} is a SKU level product dataset that consisting of 2019 product SKUs. Because there is no any image related side information provided by Materialist Product 2019, we take each image as an input query of the image search engine Pailitao~\cite{plt} and collect the title of the top1 search result as the side information. Apart from the two open datasets, we also provide two SKU level product datasets in this paper, a facial cream and mask product dataset that containing 34 thousands classes and a huge scale product dataset that containing one million classes. The two proposed datasets come from the marketplace of Alibaba. An overview of statistics about the four datasets are illustrated in Table~\ref{tab:ds}. And the long-tailed distribution of training data are shown in Fig.~\ref{fig:longtail-dist}, in which the boundaries between the head and tail classes on training data are experimentally determined as 200, 5000, 100 and 200, respectively. About the testing sets in the four datasets, each class contains approximately equal number of images for a fair evaluation.

\subsection{Evaluation metrics}\label{ssec:metric}
Considering that many classes are conceptually overlapped and ambiguous, especially in the Webvision2.0 and the SKU level product datasets, we report the results of $top1$ and $top3$ predicted labels for the evaluation of recognition performance. In addition, the evaluation of overall $top1$ and $top3$ accuracies are also conducted over the head and tail classes, respectively, to demonstrate the effect on the long tail problem.

\subsection{Implementation details}\label{ssec:imp}
In the proposed co-training scheme, we use a Resnet-50~\cite{he2016deep} initialized from a ImageNet pretrained model as the backbone convolutional neural network (CNN) in the visual stream. As shown in Fig.~\ref{fig:cotrain}, the union of the visual stream and co-training classifier in the architecture represents the baseline that carrying out a conventional classifier using image data only.
The word embeddings in the semantic stream are initialized using the $word\_embedding()$ API of Alibaba NLP toolbox (AliNLP)~\cite{alinlp}. The word embeddings of AliNLP are trained using the product titles from the marketplace of Alibaba, and well performs on many natural language processing tasks under the E-commercial circumstance at Alibaba. 

When training on such a large dataset as our proposed product dataset that containing one million classes, the size of the last fully connected~(FC) layer will be larger than the memory size of a single block of GPU~(e.g. Nvidia V100 32G). The proposed co-training system is implemented in a hybrid parallel training framework~\cite{song:2020}, in which the last FC layer is divided and sent to $M$ GPUs for distributed training. The training is carried out in a distributed computing platform of Alibaba with 60 blocks of Nvidia P100 GPUs. For a fair comparison, the baseline and the co-training approach are trained using a stochastic gradient descent~(SGD) optimizer with a same learning configuration of a batch size 256, an initial learning rate 0.1 and a step decay policy of step 1, gamma 0.8.

\subsection{Experimental results}\label{ssec:exp-ret}
\begin{table*}[htb]\centering
\begin{tabular}{|c|c|c|c|}
\hline
                                             & Baseline               & Co-training           & Gain              \\ \hline
iMaterialist Product 2019            & 57.29 (83.77)     & 58.30 (85.45)    & \textbf{1.01} (\textbf{1.68})   \\ \hline
Webvision2.0 Selected 1k             & 62.68 (79.02)     & 65.95 (80.93)    & \textbf{3.27} (\textbf{1.91})    \\ \hline
Alibaba 34K SKU Product              & 51.60 (77.04)     & 53.47 (78.89)    & \textbf{1.87} (\textbf{1.85})    \\ \hline
Alibaba 1M SKU Product               &  88.25 (97.13)    & 90.25 (97.99)    & \textbf{2.00} (\textbf{0.86})   \\ \hline
\end{tabular}
\caption{The $top1$ and $top3$ (in parentheses) classification accuracies and performance improvements compared to the baselines on the four datasets.}
\label{tab:acc-ret}
\end{table*}

\subsubsection{Overall classification accuracy}\label{sssec:comp}
As illustrated in Table~\ref{tab:acc-ret}, we report the $top1$ and $top3$ (enclosed in parentheses) classification accuracies on the four datasets. It is clearly observed that our proposed co-training approach using the side information of image titles shows performance improvements against the baselines with significant $top1$ (and $top3$) accuracies gain by 1.01\%(1.68\%) in iMaterialist Product 2019, 3.27\%(1.91\%) in Webvision2.0 Selected 1k, 1.87\%(1.85\%) in Alibaba 34K SKU Product and 2.00\% (0.86\%) in Alibaba 1M SKU Product. When the number of classes ranging from 1 thousands to 1 millions, our approach achieves consistent and significant accuracy gains all the time. The consistent improvement of classification performance demonstrates an attractive scalability for setting out practical visual recognition applications.

\subsubsection{Effect of the bilinear word attention based semantic embedding}\label{sssec:longtail}
\begin{figure*}[htb]
\begin{minipage}[b]{.99\linewidth}
  \centering
  \centerline{\includegraphics[width=12cm]{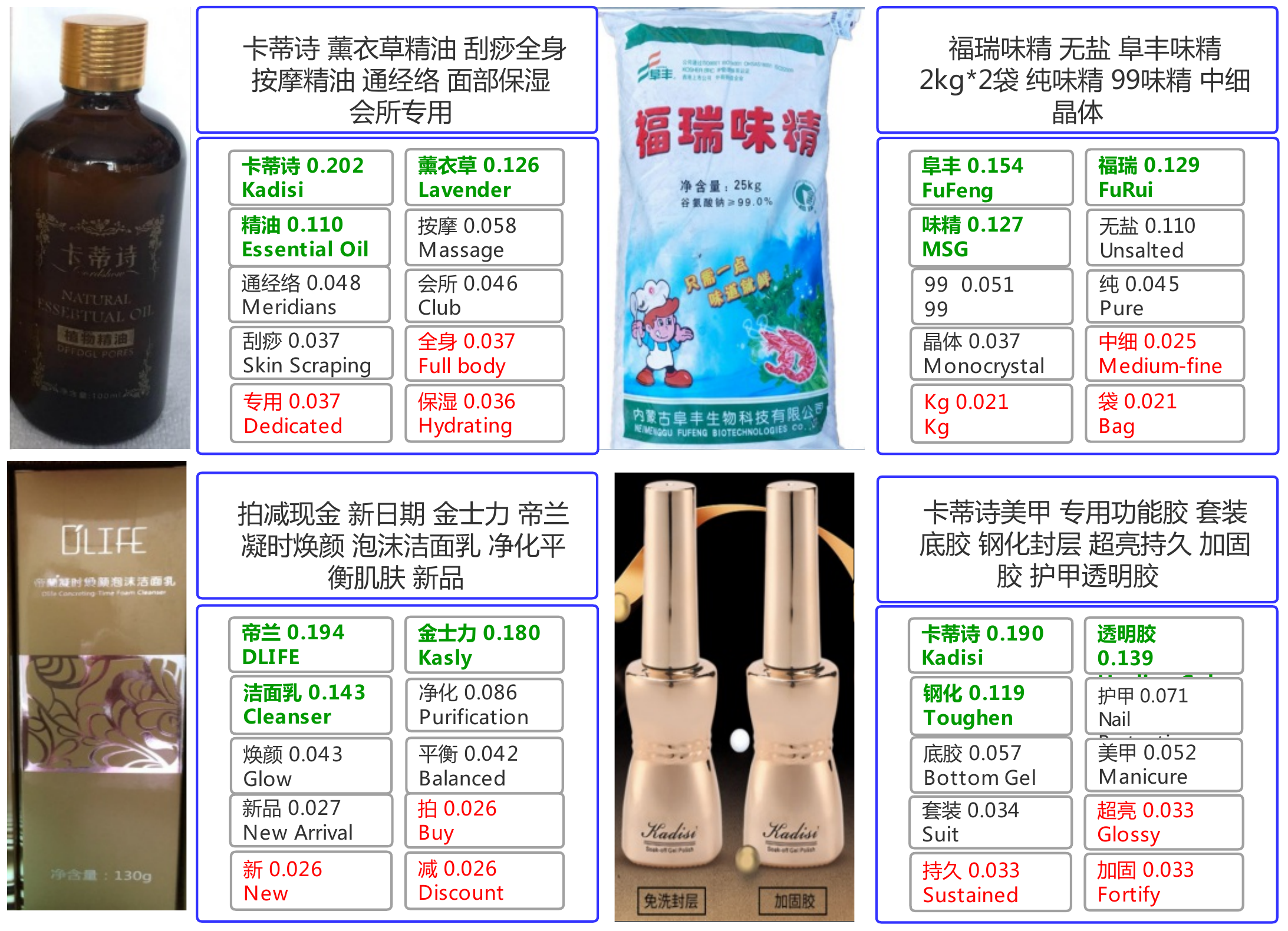}}
\end{minipage}
\caption{Visualization of learned attention word vectors on four examples. Each example consists of a image, an original image title in Chinese and a series of words with corresponding attention value in both Chinese and English. These words are displayed in the descending order of the corresponding value in the attention vector. In each example, three most important words are highlighted in green and three most unimportant words are highlighted in red.}
\label{fig:uce-exp}
\end{figure*}

For a qualitative evaluation of the proposed bilinear word attention based semantic embedding, as illustrated in Fig.~\ref{fig:uce-exp}, we visualize the learned word attention vectors of several title samples. The titles of the four examples are demonstrated in original Chinese and translated English at the same time. These words are displayed in the descending order of the corresponding value in the attention
vector. In each example, three most important words are highlighted in green and three most unimportant words are highlighted in red. It is observed that the words with larger attention weights generally refer to product names and brand names, and the words with smaller attention weights are often less helpful to distinguish the product from other products. For example, the promotion words of "discount" and "new arrival" are irrelevant to identifying a product. 

Meanwhile, as shown in Tab.~\ref{tab:biatt-ret}, we compare our proposed bilinear word attention based embedding with several other methods on the proposed Alibaba 1M SKU product dataset. The method of Alinlp embeddings mean pooling simply takes the mean of the Alinlp word embeddings as the final title embedding. The bi-LSTM method takes advantage of a bidirectional long short-term memory~(LSTM) module to generate a final title embedding. The bi-LSTM attention method introduces a self attention mechanism into the bi-LSTM method. It is observed that our proposed bilinear word attention based embedding outperforms the other four methods. The two bi-LSTM based methods take the order of words into consideration by using a sequential model bi-LSTM to describe the context. However, the inferior performance of the two methods indicate that the order of words is less helpful to carrying out a classification task. Compared to the Alinlp embeddings mean pooling method, our proposed attention method has a positive effect on boosting the performance of classification by using the bilinear word attention model.

\begin{table}[htb]\centering
\begin{tabular}{|c|c|c|}
\hline
Methods                       & Top1 Accuracy  & Gain\\ \hline
baseline                      &  88.25      & \--                          \\ \hline
Alinlp embeddings mean pooling&  89.14      & \textbf{0.89}    \\ \hline
bi-LSTM                       &  89.13      & \textbf{0.88}    \\ \hline
bi-LSTM attention             &  88.62      & \textbf{0.37}    \\ \hline
bilinear word attention       &  89.41      & \textbf{1.16}    \\ \hline
\end{tabular}
\caption{Comparison of our proposed bilinear word attention based embedding with other four methods on the proposed Alibaba 1M SKU Product dataset. }
\label{tab:biatt-ret}
\end{table}

\begin{table*}[htb]\centering
\begin{tabular}{|l|c|c|c|c|l|}
\hline
\multicolumn{1}{|c|}{} & \multicolumn{1}{l|}{} & \multicolumn{1}{l|}{\#Training Samples} & Baseline   & Co-training  & \multicolumn{1}{c|}{Gain}  \\ \hline
\multicolumn{1}{|c|}{\multirow{2}{*}{iMaterialist Product 2019}} & Head & 200 - max & 57.38 (83.49) & 58.37 (85.08) & \multicolumn{1}{c|}{\textbf{0.99} (\textbf{1.59})} \\ \cline{2-6}
\multicolumn{1}{|c|}{}                                           & Tail & 1 - 200   & 37.21 (81.40) & 41.86 (83.72) & \multicolumn{1}{c|}{\textbf{4.56} (\textbf{2.32})} \\ \hline
\multirow{2}{*}{Webvision2.0 Selected 1k}                        & Head & 5000 - max & 55.18 (72.91) & 55.86 (72.55) & \multicolumn{1}{c|}{\textbf{0.68} (\textcolor{red}{\textbf{-0.36}})} \\ \cline{2-6}
                                                                 & Tail & 1 - 5000   & 63.04 (79.26) & 66.23 (81.11) & \textbf{3.19} (\textbf{1.85})  \\ \hline
\multirow{2}{*}{Alibaba 34K SKU Product}                         & Head & 100 - max  & 62.98 (84.94) & 63.54 (85.51) & \textbf{0.56} (\textbf{0.57})  \\ \cline{2-6}
                                                                 & Tail & 1 - 100    & 49.24 (75.40) & 51.38 (77.51) & \textbf{2.14} (\textbf{2.11})  \\ \hline
\multirow{2}{*}{Alibaba 1M SKU Product}                          & Head & 200 - max  & 95.09 (99.53) & 95.61 (99.62) & \textbf{0.52} (\textbf{0.09})  \\ \cline{2-6}
                                                                 & Tail & 1 - 200    & 86.87 (96.64) & 89.16 (97.66) & \textbf{2.29} (\textbf{1.02})  \\ \hline
\end{tabular}
\caption{ The averaged $top1$ and $top3$ (in parentheses) classification accuracies and performance improvements on the head classes and the tail classes, repetitively.}
\label{tab:longtail-ret}
\end{table*}
 
\subsubsection{Effect on long-tailed distribution of training data}\label{sssec:longtail}
As shown in Table~\ref{tab:longtail-ret}, we illustrate the effect of our proposed co-training scheme on the problem of long-tailed distributed training data. Compared with the baseline on the four datasets, we report the averaged top1 and top3 classification accuracies of the head and tail classes, respectively. It is observed that our approach achieves improvement of the top1 and top3 classification accuracies in both the head classes and the tail classes. Noted that the performance gains in the head classes are more significant than the gains in the tail classes.
This phenomenon indicates that our approach improves performance of the tail classes while does not harm the performance of the head classes. In fact, our approach often can slightly improve the performance of the head classes at the same time. This mainly owes to that the procedure of knowledge transfer is bidirectional in the proposed co-training system, and the knowledge extracted from the tail classes is also beneficial to the training of the head classes.

\subsubsection{Application in Pailitao}\label{sssec:longtail}
Pailitao~\cite{plt} is a visual search application which aims to assisting customers to find the same or similar products by a mobile phone camera shot image. Pailitao is still experiencing swift growth of daily active users~(DAU). In Pailitao, the unique visitor (UV) conversion rate is a common measurement which is calculated as Equ.~\ref{equ:uv}.
\begin{align}
  \mathrm{UV\ conversion\ rate} = \frac{\mathrm{number\ of\ trading\ UV}}{\mathrm{number\ of\ visiting\ UV}}
  \label{equ:uv}
\end{align}
To further improve the UV conversion rate of Pailitao, a huge scale SKU level product recognition service is settled in Pailitao as an upgrade.
As shown in Fig.~\ref{fig:demo}, the recognition result of a query is displayed at the top panel of the page in conjunction with the results of visual search. The panel includes a clickable image, a short title and several tags of the recognized product. The product recognition service is constructed by using the proposed side information based co-training system. Compared to the original version of Pailitao, there is a relative $3.1$ percent gain of daily UV conversion rate after this upgrade.

\section{Conclusion}\label{sec:conclusion}
The phenomenon of long-tailed imbalanced training data naturally occurs under the E-commercial circumstance and challenges the performance of a large scale visual recognition task. In this paper we address the problem of long-tailed distributed training data by exploring a side information
based visual recognition co-training (SICoT) system. We firstly introduce a bilinear word attention sub-network to distinguish important words from the noisy side information, followed by generating a semantic embedding as the distilled knowledge of the side information. An end-to-end visual
feature and semantic embedding co-training system is then proposed to help transfer knowledge from the head classes to the tail classes. Experimental results on four large scale datasets demonstrate the effectiveness of the proposed approach. Our approach improves the performance of the tail classes without any harm to the head classes. Moreover, a SICoT driven product visual recognition service is settled in Pailitao and achieves a significant gain of unique visitor conversion rate. Our approach has shown good scalability ranging from medium to large scale datasets, and it is of great value for establishing industrial visual recognition applications.

\bibliographystyle{ACM-Reference-Format}
\bibliography{ms}

\end{document}